# Robust Open-Vocabulary Translation from Visual Text Representations

**Elizabeth Salesky**[1] and **David Etter**[2] and **Matt Post**[1,2]
[1]Center for Language and Speech Processing
[2]Human Language Technology Center of Excellence
Johns Hopkins University
Baltimore, Maryland, USA

## Abstract

Machine translation models have discrete vocabularies and commonly use subword segmentation techniques to achieve an 'open vocabulary.' This approach relies on consistent and correct underlying unicode sequences, and makes models susceptible to degradation from common types of noise and variation. Motivated by the robustness of human language processing, we propose the use of *visual text representations*, which dispense with a finite set of text embeddings in favor of continuous vocabularies created by processing visually rendered text with sliding windows. We show that models using visual text representations approach or match performance of traditional text models on small and larger datasets. More importantly, models with visual embeddings demonstrate *significant robustness* to varied types of noise, achieving e.g., 25.9 BLEU on a character permuted German–English task where subword models degrade to 1.9.

## 1 Introduction

Machine translation models degrade quickly in the presence of noise, such as character swaps or misspellings (Belinkov and Bisk, 2018; Khayrallah and Koehn, 2018; Eger et al., 2019). Part of the reason for this brittleness is the reliance of MT systems on *subword segmentation* (Sennrich et al., 2016) as the solution for the open-vocabulary problem, since it can cause even minor variations in text to result in very different token sequences, needlessly fragmenting the data (Table 1). These issues can be mitigated with techniques including normalization, adding synthetic noisy training data (Vaibhav et al., 2019), or often simply moving to larger data settings. However, it is impossible to anticipate all kinds of noise in light of their combinatorics, and in any case, attempts to do so add complexity to the model training process.

| Phenomena | Word | BPE (5k) | |
|---|---|---|---|
| Vowelization | كتاب | كتاب | (1) |
| | الكِتابُ | الك · ِ · ت · اب · ُ | (5) |
| Misspelling | language | language | (1) |
| | langauge | la · ng · au · ge | (4) |
| Visually Similar Characters | really | really | (1) |
| | rea11y | re · a · 1 · 1 · y | (5) |
| Shared Character Components | 확인한다 | 확인 · 한 · 다 | (3) |
| | 확인했다 | 확인 · 했다 | (2) |

Table 1: Examples of common behavior which cause divergent representations for subword models.

Humans, in contrast, are remarkably robust to all kinds of text permutations (Rayner et al., 2006), including extremes such as "l33tspeak" (Perea et al., 2008). It stands to reason that one source of this robustness is that humans process text, not from discrete unicode representations, but *visually*, and that modeling this kind of information might yield more human-like robustness. Drawing on this, we propose to translate from *visual text representations*. Our model still consumes text, but instead of creating an embedding matrix from subword tokens, we render the raw, unsegmented text as images, divide it into overlapping slices, and produce representations using techniques from optical character recognition (OCR). The rest of the architecture remains unchanged. These models therefore contain both visual and distributional information about the input, allowing them to potentially provide robust representations of the input even in the presence of various kinds of noise.

After presenting the visual text embedder (Section 2), we demonstrate the potential of visual representations for machine translation across a range of languages, scripts, and training data sizes (Section 4). We then look at a variety of types of noise, and show significant improvements in model robustness with visual text models (Section 5).

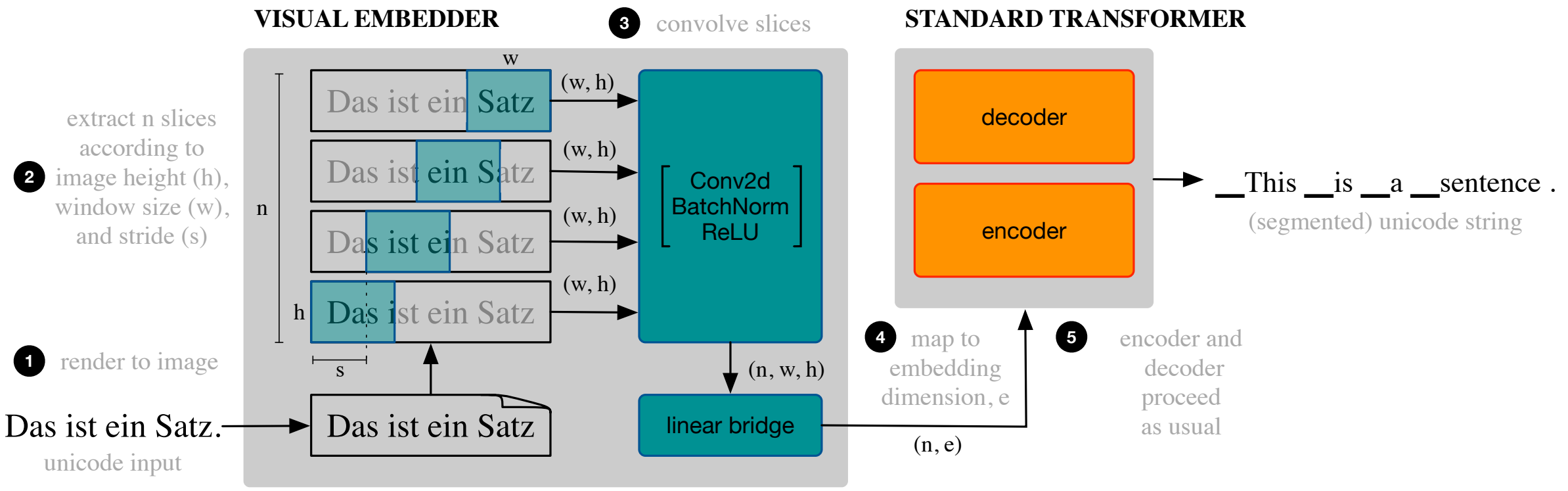

Figure 1: Visual text architecture combines network components from OCR and NMT, trained end-to-end.

# 2 Visual Text Embedder

## 2.1 Rendering text as images

Our architecture is summarized in Figure 1. The first step is to transform text into an image. We render the raw text of each input sentence into a grayscale (single color channel) image; no subword processing is used at all. The image height $h$ is a function of the maximum height of the characters given the font and font size, while the image width $w$ is variable based on the font and sentence length. We extract slices using sliding windows, similar to feature extraction for speech processing. Each window is of a specified length $w$ and full height $h$, extracted at intervals $s$ determined by a set stride. We experimentally tune each of these parameters per language pair (see Section 3.4).

## 2.2 Visual representations

The slices output from the rendering stage are analogous to subword text tokens. The next step produces "embeddings" from these slices. Embeddings typically refer to entries in a fixed size weight matrix, with the vocabulary ID as an index. Our image slices are not drawn from a predetermined set, so we cannot work with normal embeddings. Instead, we use the outputs of 2D convolutional blocks run over the image slices, projected to the model hidden size, as a continuous vocabulary.

While OCR models for tasks such as handwriting recognition require depth that impacts training and inference speed, our task differs significantly. OCR tasks contend with varied image backgrounds, varied horizontal spacing, and varied character 'fonts,' sizes, colors, and saliency. Visually rendered text is uniform along each of these characteristics by construction. Accordingly, we can use simpler image processing and model architectures without performance impact.

Our core experiments use a single convolutional block ($c = 1$) followed by a linear projection to produce flattened 1D representations as used by typical text-to-text Transformer models, but here the representations are drawn from a continuous space rather than a predetermined number of embeddings. A convolutional block comprises three pieces: a 2D convolution followed by 2D batch normalization and a ReLU layer. The 2D convolution uses only one color channel, and padding of 1, kernel size of 3, and stride of 1, which results in no change in dimensions between the block inputs and outputs. We contrast the $c = 1$ setting with two others: $c = 0$ and $c = 7$. When $c = 0$, the model is akin to the Vision Transformer (Dosovitskiy et al., 2021) from image classification where attentional layers are applied directly to image slices[1] after a flattening linear transformation. With $c = 7$, we compare the depth of VistaOCR (Rawls et al., 2017), a competitive OCR model, but without its additional color channels and subsequent max-pooling.[2]

After replacing text embeddings with visual representations, the standard MT architecture remains the same. The full model is trained end-to-end with the typical cross-entropy objective. All models are trained using a modified version of `fairseq` (Ott et al., 2019), which we release with the paper.[3]

# 3 Experimental Setup

## 3.1 Training data

We experiment with two data scenarios, a *small* one (MTTT) and a *larger* one (WMT).

[1] Unlike ViT, we extract overlapping slices with the full image height, rather than smaller non-overlapping square slices.
[2] VistaOCR iteratively grows the channel axis from 3 color channels to 256 and adds 2 interleaved max-pooling layers.
[3] `https://github.com/esalesky/visrep`

**MTTT.** We use the MTTT dataset to compare traditional text models with visual text models across a range of languages and scripts, using similarly sized data. We use the Multitarget TED Talks Task (MTTT), a collection of TED[4] datasets with ∼200k training sentences (Duh, 2018). Specifically, we use the data for the Arabic (ar), Chinese (zh), Japanese (ja), Korean (ko), Russian (ru), French (fr), and German (de) to English (en) tasks.

**WMT.** We also experiment with two larger datasets derived from the 2020 shared task in news translation from the Conference on Machine Translation (WMT20). For German–English, we use all provided data except Paracrawl and Commoncrawl. We filter out sentence pairs that don't match on language ID as reported by `fasttext` (Joulin et al., 2016b,a), pairs with a raw length ratio of more than 3 to 1, pairs where raw source or target length is greater than 100, and all duplicate pairs, leaving 4.9M sentence pairs. We train a joint unigram SentencePiece model of size 10k.

For Chinese, we use all provided data except UNv1.0 and the backtranslations. We filter in the same way, except that we do not apply ratio filtering. This yields 8.7M sentence pairs. We train separate source and target unigram SentencePiece models of sizes 20k and 10k, respectively. More details can be found in Table 13 in the Appendix.

### 3.2 Test sets

**MTTT.** Our main results are on the 1,982 segment multi-way parallel MTTT test sets.

**MTNT.** To evaluate model robustness on data with naturally occurring noise, we use the Machine Translation of Noisy Text (MTNT) test sets (Michel and Neubig, 2018). The MTNT test sets used were created from comments from Reddit in French, German, and Japanese which have been professionally translated from English. By virtue of their domain, these test sets contain "noisy" text with natural typos, semantic use of visually similar characters, abbreviations, grammatical errors, emojis, and more. MTNT has recently been used for evaluation in the WMT'19 and '20 Robustness tasks (Li et al., 2019; Specia et al., 2020).

**WIPO.** We use the World Intellectual Property Organization (WIPO) COPPA-V2 corpus (Junczys-Dowmunt et al., 2016) to evaluate robustness on data with naturally occurring noise for Russian-English. The WIPO corpus consists of parallel sentences from international patent application abstracts.

[4]`https://www.ted.com`

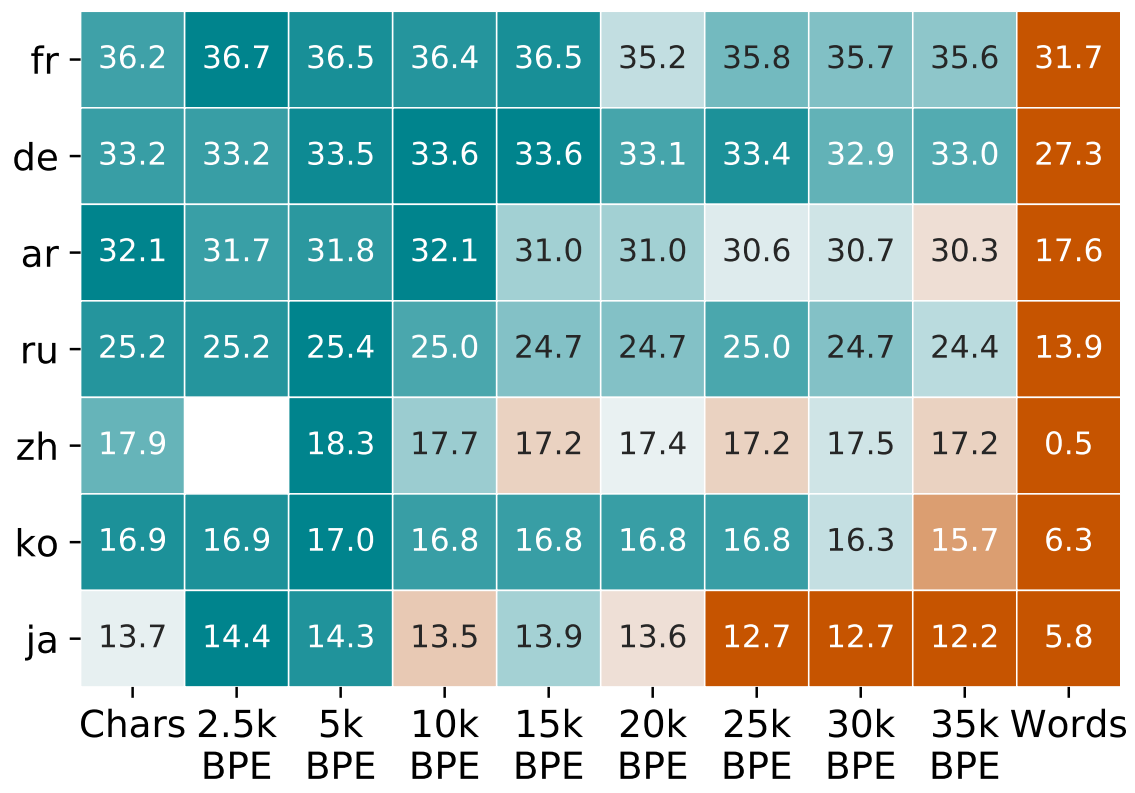

| | Chars | 2.5k BPE | 5k BPE | 10k BPE | 15k BPE | 20k BPE | 25k BPE | 30k BPE | 35k BPE | Words |
|---|---|---|---|---|---|---|---|---|---|---|
| fr | 36.2 | 36.7 | 36.5 | 36.4 | 36.5 | 35.2 | 35.8 | 35.7 | 35.6 | 31.7 |
| de | 33.2 | 33.2 | 33.5 | 33.6 | 33.6 | 33.1 | 33.4 | 32.9 | 33.0 | 27.3 |
| ar | 32.1 | 31.7 | 31.8 | 32.1 | 31.0 | 31.0 | 30.6 | 30.7 | 30.3 | 17.6 |
| ru | 25.2 | 25.2 | 25.4 | 25.0 | 24.7 | 24.7 | 25.0 | 24.7 | 24.4 | 13.9 |
| zh | 17.9 | | 18.3 | 17.7 | 17.2 | 17.4 | 17.2 | 17.5 | 17.2 | 0.5 |
| ko | 16.9 | 16.9 | 17.0 | 16.8 | 16.8 | 16.8 | 16.8 | 16.3 | 15.7 | 6.3 |
| ja | 13.7 | 14.4 | 14.3 | 13.5 | 13.9 | 13.6 | 12.7 | 12.7 | 12.2 | 5.8 |

Figure 2: Baseline results on MTTT TED across BPE segmentations with optimized batch size.

**WMT.** For the larger data experiments from German and Chinese, we report results on the WMT'20 newstest sets (Barrault et al., 2020).

### 3.3 Baseline text models

All baseline text models are trained using `fairseq`. For our 7 language pairs from the MTTT TED dataset, we follow the recommended `fairseq` architecture and optimization parameters for IWSLT'14 de-en which is of the same size and domain: 6 layers each for encoder and decoder, with 4 attention heads per layer, with slight modifications to batch size, vocabulary, and label smoothing $p = 0.2$.

We tune the subword vocabulary for each language pair and dataset. We saw no difference between joint/disjoint vocabularies, so use separate vocabularies to create a direct comparison with the visual text models: the same target vocabulary is used for both and only the source representations are varied. We tuned ∼5k BPE intervals from 2.5k–35k[5] to optimize source language BPE granularity with the target vocabulary held constant at 10k BPE. We additionally compare character-level and word-level models; to produce word-level segmentations for Chinese, we use `jieba`,[6] and for Japanese, we use `kytea` (Neubig et al., 2011). The character vocabulary for Chinese is greater than 2.5k so we do not have a BPE model of this size. Our best performing BPE models used source vocabularies of approximately 5k (see Figure 2).

We jointly tuned batch size and subword vo-cabularies for each language pair and found significant (1–15 BLEU) improvements with a larger batch of 16k tokens over the suggested 4096, particularly for Chinese, Japanese, and Korean. Our baselines improve ∼2 BLEU over previous work on the MTTT dataset (Shapiro and Duh, 2018).

[5]For the MTTT datasets, ∼40k BPE recovers words.

[6]`https://github.com/fxsjy/jieba`

| Lang | Text | | Visual text | | | |
|---|---|---|---|---|---|---|
| | BPE | char | $s = 5$ | $s = 10$ | $s = 15$ | $s = 20$ |
| ar | 24.4 | 78.9 | 97.1 | 48.8 | 32.7 | 24.6 |
| de | 32.3 | 104.3 | 130.5 | 65.5 | 43.8 | 33.0 |
| fr | 28.8 | 107.6 | 130.2 | 65.4 | 43.7 | 32.9 |
| ja | 22.5 | 36.9 | 95.5 | 48.0 | 32.1 | 24.2 |
| ko | 24.7 | 50.8 | 97.0 | 48.7 | 32.6 | 24.6 |
| ru | 27.1 | 94.7 | 132.7 | 66.6 | 44.5 | 33.5 |
| zh | 23.0 | 29.8 | 75.6 | 38.1 | 25.5 | 19.3 |
| **Time** | 1.0× | 2.3× | 3.9× | 2.0× | 1.4× | 1.2× |

Table 2: Average sequence lengths of MTTT data for text models and visual models with varying stride $s$. The bottom row shows training time relative to the fastest model (BPE) with $c = 1$.

| | Text | Visual Text | | |
|---|---|---|---|---|
| | | $c = 1$ | $c = 0$ | $c = 7$ |
| ar-en | 32.1 | 31.6 | 30.4 | 30.2 |
| de-en | 33.6 | 35.1 | 34.0 | 34.3 |
| fr-en | 36.7 | 36.2 | 36.0 | 35.3 |
| ja-en | 14.4 | 13.1 | 11.2 | 12.8 |
| ko-en | 17.0 | 16.6 | 15.2 | 16.2 |
| ru-en | 25.4 | 25.0 | 23.4 | 23.3 |
| zh-en | 18.3 | 17.6 | 16.5 | 17.0 |

Table 3: BLEU scores on MTTT test sets for models trained on the MTTT data. The number of convolutional blocks is denoted by $c$. Our best visual models ($c = 1$) approach parity with optimized text baselines.

For the larger data settings, we train Transformer base models[7] with dropout 0.1 and learning rate 4e-4. We use a batch size of 40k tokens, and train until held-out validation fails to improve for ten epochs. For German, we use a shared unigram subword vocabulary of size 10k. For Chinese, we train separate models of size 20k and 10k, respectively. No other preprocessing was used.

### 3.4 Visual text models

Our visual text models replace the source embedding matrix in the text models with the visual text embedder from Section 2. The model architecture otherwise remains unchanged: we use the same Transformer settings, and the target language vocabulary is 10k BPE. We experiment with parameters for the visual text embedder to find which are significant for this new task in Section 4, with hyperparameter sweeps in Appendix A.

We use the `pygame` Python package[8] with the Google Noto font family[9] to render text. For Latin and Cyrillic scripts, we use NotoSans; for Arabic, NotoNaskhArabic; and for the ideographic languages, NotoSansCJK JP. No preprocessing is applied before rendering.

While our visual text models remove the source embedding matrix, they may add parameters from convolution blocks if used to compute representations. Our best models typically reduce the number of model parameters, and in the worst case increase overall parameters by 1% (from 36.7M to 36.9M), determined by window size and number of convolutional blocks. Computation time increases compared to BPE due to longer source sequences, but our best performing models are faster (with shorter sequences) than character models (Table 2). Time to render text during inference is negligible—comparable to subword segmentation at fractions of a second.

[7] 6 layers, 8 attention heads, embed dim 512, FF dim 2048
[8] `https://www.pygame.org`
[9] `https://www.google.com/get/noto`

## 4 Chasing Translation Parity

State-of-the-art translation models use subword vocabularies, which yield best performance when tuned per language pair and task (Salesky et al., 2018; Ding et al., 2019). Our visual text approach avoids predetermining a fixed model vocabulary. On the one hand, this allows us to represent even unanticipated characters; on the other, optimizing a finite model vocabulary per task may improve performance. Our first question, therefore, is whether visual text can recover scores produced by baselines with optimized subword vocabularies.

On the smaller MTTT dataset, we can nearly recover the best results from the most optimal BPE segmentation *without* explicit input segmentation, solely from visual representations with a sliding window. Table 3 compares our best visual text models to our best text baselines on MTTT. The best visual text results use $c = 1$ convolutional block, which adds some structural biases from convolutions without excessive visual depth. We show $c = 0$ and $c = 7$ for comparison, which represent no convolutional blocks and the depth of recent state-of-the-art OCR models, respectively. We find greater visual capacity through a larger number of convolutional blocks does not improve results for our task. Increased convolutional depth also comes at a cost: compared to $c = 1$, $c = 7$ adds 2.6M additional parameters and 5× longer training time. In this setting, $c = 0$ is consistently below $c = 1$. Our analysis focuses on $c = 1$.

| | Text | Visual Text | | |
|---|---|---|---|---|
| | | $c = 0$ | $c = 1$ | $c = 7$ |
| de-en | 33.9 | 32.9 | 32.5 | — |
| zh-en | 20.2 | 21.3 | 20.5 | — |

Table 4: BLEU scores on WMT test sets with the larger WMT models.

| DE-EN | $c = 1,\ font = 8pt$ | | | |
|---|---|---|---|---|
| stride↓/window→ | 15 | 20 | 25 | 30 |
| 10 | 33.4 | 33.1 | 33.3 | 33.6 |
| 15 | 33.9 | 32.9 | 31.3 | 32.9 |
| 20 | | 32.0 | 30.3 | 32.4 |
| 25 | | | 30.4 | 30.9 |

Table 5: German–English BLEU scores on MTTT, tuning stride and window length with fixed batch size.

On our larger data scenarios, we see our best visual text models approach (de-en) or exceed (zh-en) the text-based baselines (Table 4). This suggests our approach scales and its efficacy is not limited to lower-resource settings. With more data, $c = 0$ slightly outperforms the $c = 1$ model, suggesting this 'direct' model may simply require more training data.

As a new approach, it is not known from the outset which hyperparameters for visual representations may affect performance. We conducted experiments to determine significant hyperparameters and best parameter ranges for visual text experiments: namely, for window length, stride, font size, batch size, and CNN kernel size. We see similar hyperparameter trends across language pairs. We find font size is not significant as long as it is sufficiently large to not affect image resolution for more visually dense scripts (at least 10pt—see Table 12 in the Appendix), and CNN kernel size of 3 × 3 and batch size of $20k$ to be consistently best. We always use a window length greater than or equal to stride length so that no text is dropped. Table 5 shows varied window length and stride values for de-en; additional language pairs and parameter combinations can be found in Appendix A. As stride length increases (creating less overlap between windows) performance typically decreases: our best results typically use stride 10. Optimal window length exhibited the biggest difference between languages. We show ablation experiments isolating the role of sliding window segmentation in Appendix B.

# 5 Robustness to Noise

We hypothesize that without a fixed vocabulary and with associations between visually similar character spans, our visual text models will be more robust to noise than text-based representations, where noise causes divergent subword representations (see Table 1 for motivating examples). To test this, we evaluate on two different settings: induced synthetic noise, and naturally occurring noise from sources such as Reddit. Synthetic noise allows us to test various settings for all language pairs, while natural noise is limited by dataset availability. Examples of induced noise, and the resulting model inputs and outputs for both text and visual text models, can be found in Table 6.

## 5.1 Synthetic noise

Inducing noise enables us to control the type and frequency with which noise occurs. We compare two types of synthetic noise: **visually similar characters** (e.g., l33tspeak, unicode codepoints which are visually similar) and **character permutations** (e.g., Cmabrigde). For all synthetic noise experiments, we induce noise at the token-level on the source side of our baseline dataset, MTTT TED. Each token may be noised with probability $p$ from $p = 0.1$ to $1.0$ by intervals of $0.1$.

**Visually similar characters.** Different unicode characters may share visually similar characteristics. Such characters may be substituted intentionally, such as in l33tspeak where characters such as numbers are used in place of visually similar Roman alphabet letters, or unintentionally, where characters from another script appear in place of the expected unicode codepoints for a given language and script due to e.g., use of multiple keyboards or OCR errors (Rijhwani et al., 2020). For some languages without a unicode standard, multiple unicode sequences which render the same are all in common use (e.g., Pashto). As shown in Figure 3, such errors can be very inconspicuous.

| | |
|---|---|
| Arabic–English | |
| src | أنا كندية، وأنا أصغر أخواني السبعة |
| `diacritics` 1.0 | أَنا كَنَدِيَّةٍ ، وَأَنا أَصْغَرِ إِخْوانِي السَبْعَةِ |
| ref | I'm Canadian, and I'm the youngest of seven kids. |
| in$_{vis}$ | 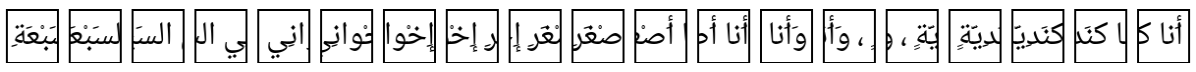 |
| out$_{vis}$ | I'm a Canadian, and I'm the youngest of my seven sisters. |
| COMET | 0.764 |
| in$_{text}$ | ـ ِأ َان ـك َن َد ِي ّ َة ـ ٍ ، ـو َ ـأ َان ـ أ َص ْغ َر ِ ـإ ِخ ْو ا ن ِي ـسلا َب ْع َ ـة |
| out$_{text}$ | We grew up as a teacher, and we gave me a hug. |
| COMET | -1.387 |
| French–English | |
| src | Un homme de 70 ans qui voudrait une nouvelle hanche, pour qu'il puisse retourner au golf ou s'occuper de son jardin. |
| `l33tspeak` 0.1 | Un homme de 70 an5 qu1 voudrait un3 nouvelle h4nche, pour qu'il pui5s3 re7ourner au golf ou s'occuper de son jardin. |
| ref | Some 70-year-old who wanted his new hip so he could be back golfing, or gardening. |
| in$_{vis}$ | 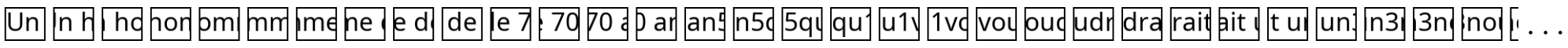 |
| out$_{vis}$ | A 70-year-old man who would like a new hip, so that he could turn to golf or take care of his garden. |
| COMET | 0.564 |
| in$_{text}$ | _Un _homme _de _70 _an 5 _qu 1 _voudr ait _un 3 _nouvelle _h 4 nch e , _pour _qu ' il _pu i 5 s 3 _re 7 our ner _au ... |
| out$_{text}$ | A 75-year-old man wants a third new hip, so that he can punish himself for the golf or take care of his garden. |
| COMET | 0.091 |
| German–English | |
| src | Aber Sie müssen zuerst zwei Dinge über mich wissen. |
| `swap` 0.5 | Abre Sie müssen zuerts wzei Dnige über mcih wisse.n |
| ref | But first you need to know two things about me. |
| in$_{vis}$ | 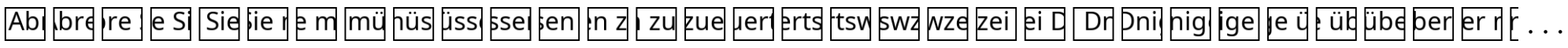 |
| out$_{vis}$ | But you have to know two things about me first. |
| COMET | 0.897 |
| in$_{text}$ | _Ab re _Sie _müssen _zu ert s _w z ei _D n ige _über _m ci h _ wiss e . n |
| out$_{text}$ | But you've got to get into a little about you. |
| COMET | -0.520 |
| Korean–English | |
| src | 전여전히제아픈골반으로지탱하고있었겠죠. 그건정말실망스러웠죠. |
| `cambridge` 0.3 | 전여전히제아픈골으반로지탱하고있었겠죠. 그건정말실망스러웠죠. |
| ref | I would still be on my bad hip. That was so disappointing. |
| in$_{vis}$ | 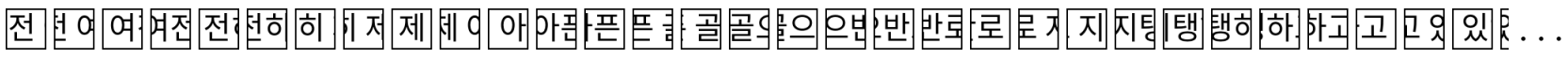 |
| out$_{vis}$ | I was still supported by my sick bone, which was really disappointing. |
| COMET | 0.198 |
| in$_{text}$ | _ 전 _ 여전히 _ 제 _ 아픈 _ 골으반로 _ 지탱하고 _ 있었겠죠. _ 그건 _ 정말 _ 실망스러웠죠. |
| out$_{text}$ | I was still living in my sick celeste, and it was quite disappointing. |
| COMET | -0.087 |
| Russian–English | |
| src | Я расскажу вам об этой технологии. |
| `unicode` 0.8 | R расскажу ваM об этой технологии |
| ref | I'm going to tell you about that technology. |
| in$_{vis}$ | 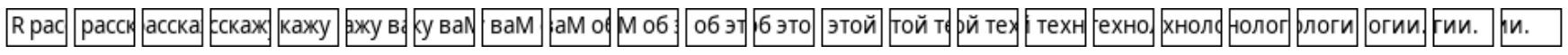 |
| out$_{vis}$ | I'm going to tell you about this technology. |
| COMET | 0.923 |
| in$_{text}$ | _R _р а с с к а жу _в а M _о б _э т о й _т е х н о л о ги и . |
| out$_{text}$ | I'm going to put my mouth in the dam of ecsta chhallogi. |
| COMET | -1.236 |

Table 6: Examples of data with induced noise, and the resulting inputs and outputs for text and visual text models. One example is shown for each type of tested noise: `unicode`, `diacritics`, `l33tspeak`, `swap`, and `cambridge`. For each example, we show the original source sentence (src); noise induced with probability $p$; the reference translation (ref); rendered text with sliding windows (in$_{vis}$); visual text model output (out$_{vis}$); BPE'd text input (in$_{text}$); text model output (out$_{text}$); and COMET (Rei et al., 2020) scores computed using the default model (wmt-large-da-estimator-1719).

The invention belongs to the field of biotechnology, pharmaceutics and medicine, it could be applied for the production of drugs and for the realization of medicinal technologies, particularly for the immunotherapy of oncological diseases.

Figure 3: Different unicode codepoints may appear visually similar. In this English example from WIPO (Junczys-Dowmunt et al., 2016), all characters in red are not from the Roman alphabet but Cyrillic.

We induce noise in the form of Latin characters which are visually similar to Cyrillic characters for Russian (`unicode`), diacritization for Arabic (`diacritics`), and l33tspeak for French and German (`l33tspeak`). We use CAMeL Tools (Obeid et al., 2020) for Arabic diacritization.

Figure 4 shows that the visual text model has almost no degradation in performance with `unicode` noise, even when 100% of characters with a mapping to another visually similar unicode codepoint have been substituted. However, the text model quickly degrades towards 0 as substitutions cause mismatches with BPE vocabularies. Character-based models are similarly unable to handle OOV codepoints, and characters in extremely novel contexts, as found with this type of noise: at $p = 0.5$, our character model has a disappointing 0.2 BLEU.

The substitution of visually indistinct codepoints is perfectly suited to our approach, and it is unsurprising that it does so well. But what about noise that does produce visual variation? Visually, Arabic diacritization represents an addition of a small number of pixels (+1-5%) which generally do not affect the spatial relationship between base characters. However, at the unicode level, diacritization inserts codepoints that break up adjacent character sequences required for subword matches (see Table 1). While visual text representations are relatively robust to diacritized text, text-based models are significantly negatively impacted: Figure 4 shows decreases of at most 4 BLEU with visual text but up to 31 BLEU with BPE.

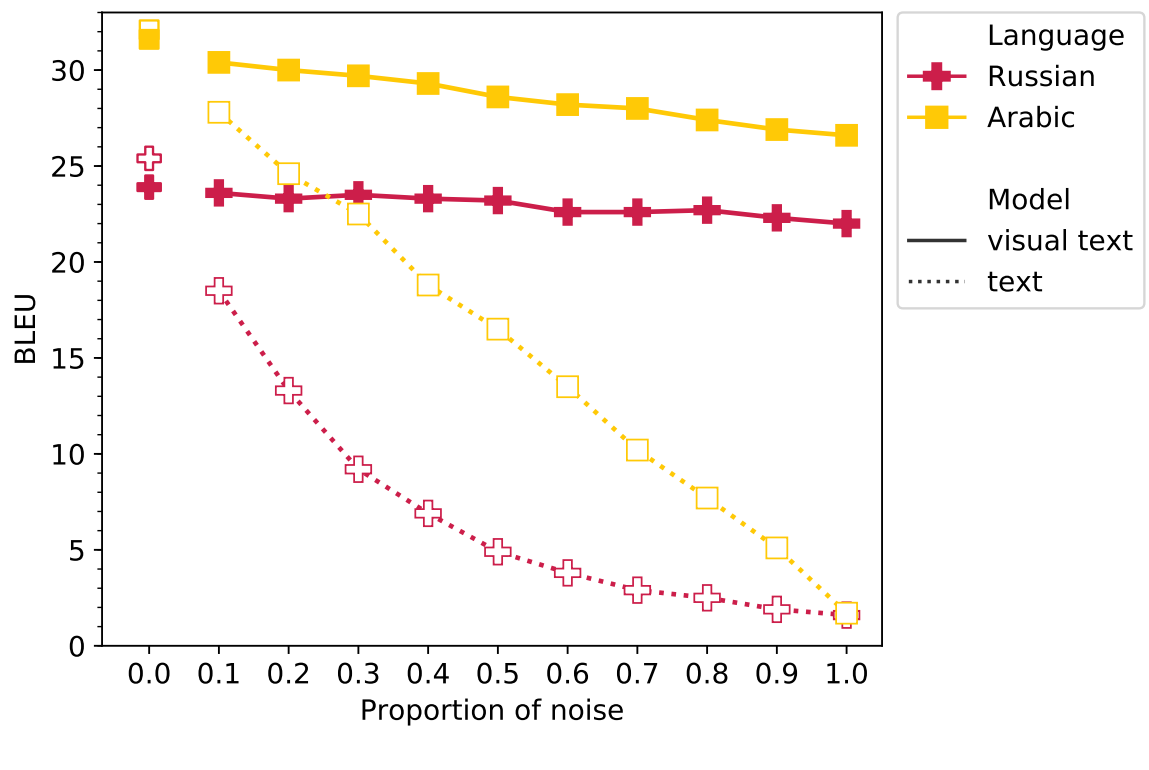

Figure 4: **Visual noise: `unicode` and `diacritics`.** Inducing visually similar codepoint differences barely affects visual text, but breaks BPE representations.

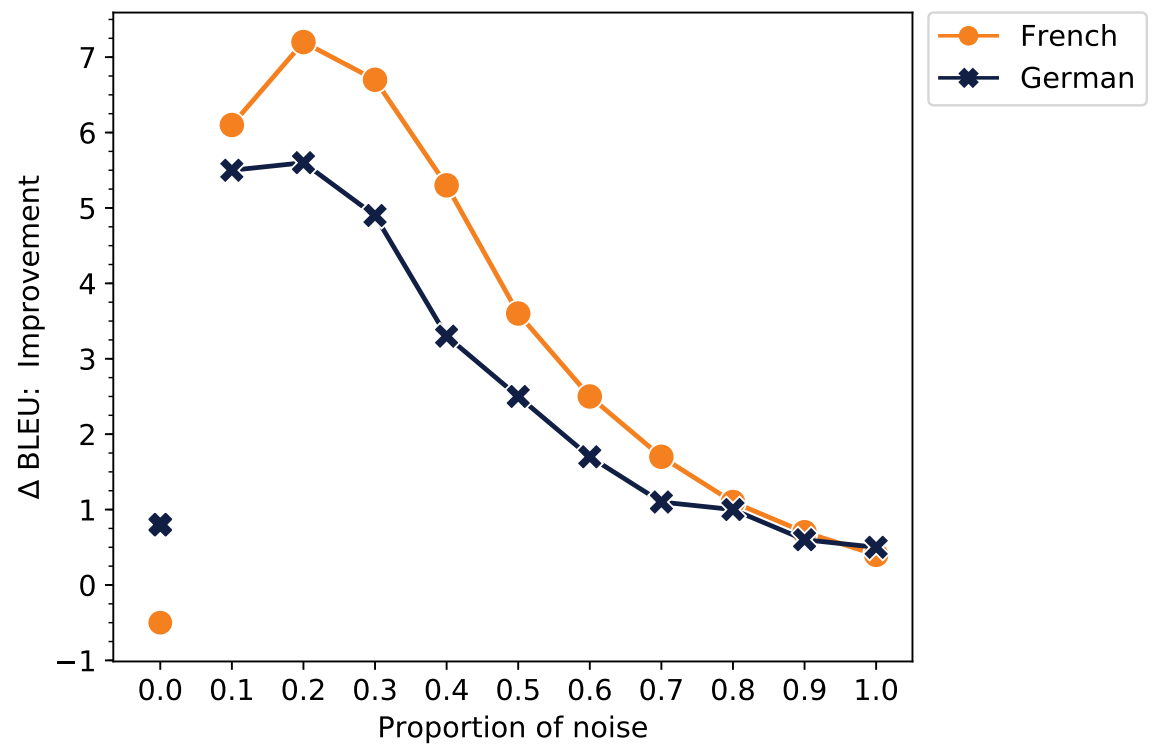

Figure 5: **Visual noise: `l33tspeak`.** ΔBLEU is shown for readability; absolute BLEU can be found in Figure 9 in Appendix D. For `l33tspeak`, improvements with visual text diminish with higher levels of noise.

Finally, we look at `l33tspeak`. Here, a reader understands from the unexpected presence of a number that a substitution has been made, and is able to form a mapping to a similar alphabetic letter. However, '4' and 'a' are not necessarily more visually similar in many fonts than say '7' and 'z'; conventional use often dictates l33tspeak substitutions moreso than visual similarity. Figure 5 shows that while both visual text models and text models are negatively affected by induced `l33tspeak`, the visual text models for both language pairs significantly outperform the text models in these conditions. With up to 30% of tokens containing `l33tspeak` mappings, the visual text models for both German and French perform >5 BLEU better than the text models.

Normalization cannot fully address these challenges for text models; see Appendix C for results.

**Character permutations** are challenging both for subword models, which necessarily back off to smaller units in the presence of OOVs (Table 1), and character-based models (Belinkov and Bisk, 2018).[10] Here we experiment with two types of synthetic noise used by Belinkov and Bisk to compare visual text models to traditional text models.

**`Swap`** : Swapping adjacent characters (e.g., *language→langauge*) is common when typing quickly. We perform one swap per selected word.

[10]If word-internal order isn't modeled (Sakaguchi et al., 2017)

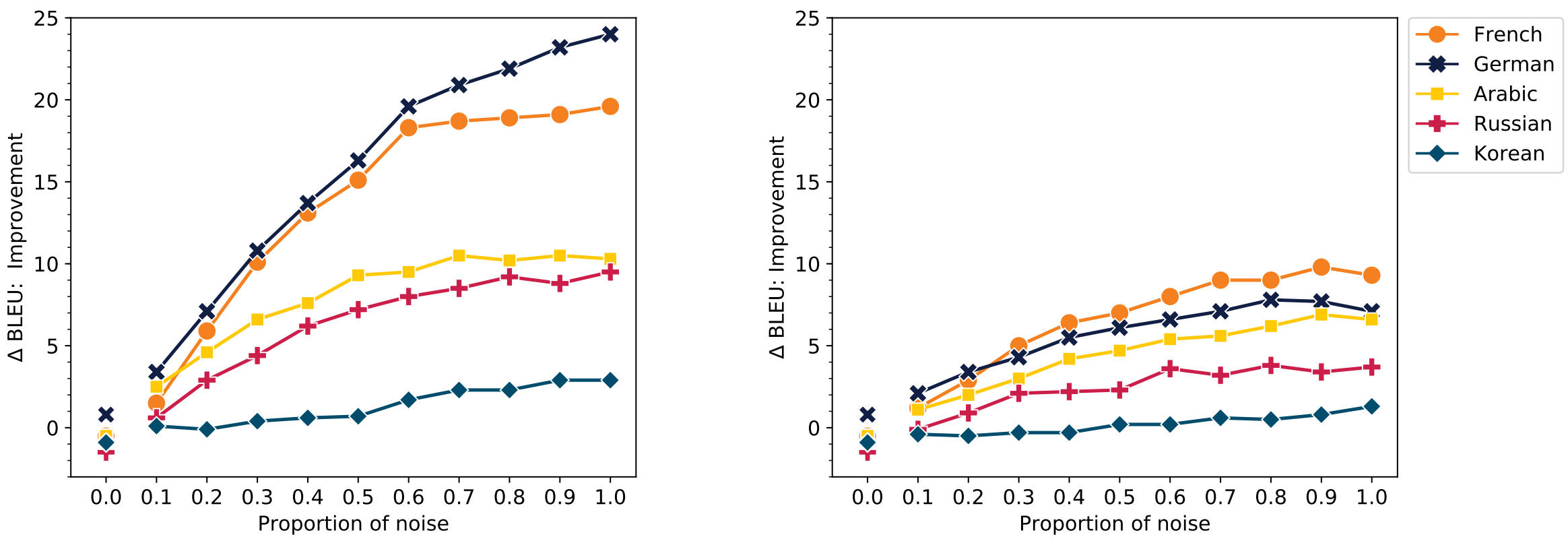

(a) `swap` (b) `cam`

Figure 6: **Character permutations: `swap` and `cam`.** Figures are shown in ΔBLEU (improvement with visual text compared to text baseline) for readability; absolute BLEU can be found in Figure 8 in Appendix D.

This noise can be applied to words of length ≥2.

**`Cam`** : The purported Cambridge spelling experiment of spam mail fame illustrates the remarkable robustness of humans to character permutations[11] when the first and last character are unchanged (e.g., *language→lnagauge*). To enable word-medial permutations, this noise can be applied to words of length ≥4.

We do not apply character permutations to Chinese or Japanese text, since most tokens contain two or fewer characters after word segmentation.

Visual text representations result in significant improvements for character permutations, particularly at higher levels of noise. Figure 6 shows the stark contrast in relative performance between the two models: though a slight gap in performance remains for some of our models on clean text, with even 10% induced noise this gap has been closed. Improvements of up to 24 BLEU on German–English concretely mean that our visual text model achieves 25.9 BLEU on a task where the subword-based model has degraded to 1.9 BLEU. Figure 7 in Appendix D shows absolute degradation in performance for each model and permutation type.

Character permutations exhibit the opposite trend of visual noise: while improvements over text models decreased as more tokens contained visual noise, for permutations, improvements strongly increased with greater levels of noise. This may be because visual noise often involves character *substitutions* rather than *permutations*. Permutations affect a greater percentage of the character sequence for a given token, which shatter subword representations. While subword models can use context to recover when only 10% of tokens contain permutations, at higher levels of noise, they cannot. When 100% of tokens contain `swaps`, for example, the German 5k BPE model backs off to 2.25× more subwords (most words are characters-only) than for non-noised text.

[11] With a cost to reading speed (McCusker et al., 1981; Rayner et al., 2006).

### 5.2 Natural noise

Natural noise—as found in informal sources, such as Reddit—contains many additional types of noise, including keyboard typos (where nearby keys are substiyuted), substitutions of phonetically similar characterz or worts, unconventional s p a c e s and repetitionsss for effect or error, natural mispelling, and noisy spans which extend beyond individual tokens, among others. Parallel text created from 'found' data (MTNT: Reddit; WIPO: patents) contains such noise in natural contexts.

| | MTNT | | | WIPO |
|---|---|---|---|---|
| **Model** | **fr-en** | **ja-en** | **de-en** | **ru-en** |
| Text, subword | 26.4 | 4.3 | 18.2 | 9.9 |
| Text, character | 26.7 | 3.7 | 20.7 | 10.3 |
| Visual text | 26.3 | 5.2 | 20.4 | 10.5 |

Table 7: Zero-shot performance on natural noise.

Table 7 compares visual text models to text models using both subword and character-level representations on MTNT and WIPO test sets. We continue to test our MTTT-trained models in a zero-shot setting, which makes domain a confounding variable for these test sets. The domain mismatch proved challenging for all models. We see that character-level models are in some cases more robust than subwords, but are unable to re-cover from the variation in others (ja-en) where the visual text model does best. The visual text models improve over subword models and perform competitively with character-level models for German–English, where we have reached parity on our clean data case (Table 3), and Russian–English, where the WIPO patent data has a significant number of unicode OCR errors (as illustrated in Figure 3) and occasional Roman alphabet characters (e.g., for chemical formulas): 3% of source characters in the Russian WIPO test set are outside the Cyrillic unicode codepoint range.

## 6 Related Work

Visual representations of text have previously been explored for other NLP tasks, primarily for Chinese, with mixed results. Liu et al. (2017) used visual representations from CNNs over rendered text in Chinese, Japanese, and Korean for text classification. Dai and Cai (2017) similarly used convolutions over character-level images for Chinese for downstream language modeling and word segmentation. Meng et al. (2019) used hybrid convolutional networks to improve several NLP tasks for Chinese, using historical scripts for additional pictographic information. Sun et al. (2018, 2019) created dense fixed-size square text renderings in Chinese and English for convolutions for downstream sentiment analysis. Ryskina et al. (2020) used visual similarity for Russian romanization.

In machine translation, visual information was also first used for Chinese. Initial work improved translation models by initializing character embeddings with linearized bitmaps of each character (Aldón Mínguez et al., 2016; Costa-jussà et al., 2017), and more recently, with linearized images compressed with PCA, which improved model robustness (Wang et al., 2020b). Nikolov et al. (2018) incorporated visual compositionality of Chinese characters for MT through Wubi ASCII encodings. Eger et al. (2019) assessed the impact of visual perturbations on current NMT systems, and augmented character-level text embeddings with visual information to make systems more robust to such attacks. Mansimov et al. (2020) explored the challenging task of image-to-image translation; while their approach is exploratory and not yet competitive, and producing images introduces difficulties with evaluation, it is similarly motivated by a desire to do away with fixed, pre-defined vocabularies and tokenization.

Previous work has explored the impact of synthetic and natural noise on neural MT (Belinkov and Bisk, 2018), and the use of character-aware word embeddings (Kim et al., 2016; Sakaguchi et al., 2017; Cherry et al., 2018; Clark et al., 2021) to increase generalizability and robustness. While research regularization and dropout techniques for BPE (Kudo, 2018; Provilkov et al., 2020) have improved model robustness, discrete vocabulary sets still creates challenge in many use cases. Recent work has also explored byte-level BPE (Radford et al., 2019; Wang et al., 2020a) to create models which are not restricted to the unicode ranges seen in training, though models using BBPE may require additional training examples.

## 7 Conclusion & Future Work

We introduced visually rendered text for continuous open-vocabulary translation. We showed that our models, trained on seven language pairs and in two data settings, approach or match the performance of traditional text models. Further, we showed that visual text models are more robust to many kinds of induced noise, including the substitution of visually similar characters and character permutations. An important benefit of our approach is that it operates on raw text, doing away with the standard preprocessing routines that include normalization, tokenization, and subword segmentation.

We believe our approach opens many avenues for future work. Standard data techniques from OCR (such as varied font and font size) and training on noise would likely further improve robustness. There are many possible visual architectures, and visual pretraining has benefited vision tasks (Dosovitskiy et al., 2021). There is nothing to preclude our approach from working on larger datasets. While effective, it is not clear that sliding window segmentation is optimal; improving segmentation could close remaining performance gaps. Since our approach does away with discrete vocabularies, visual text models could be used to transfer to new languages and scripts without requiring transliteration or normalization, or retraining models from scratch. Finally, it is appealing to consider this approach for additional tasks such as language identification (Caswell et al., 2020, Table 2) or spam detection. Any NLP task that requires robust, open-vocabulary representations could benefit from our approach.

## Acknowledgments

We thank Adi Renduchintala for early enthusiasm and proofs of concept, Paul McNamee for data pointers and examples (particularly Figure 3), and Facebook for a 2019 research award to the third author.

## References

David Aldón Mínguez, Marta Ruiz Costa-Jussà, and José Adrián Rodríguez Fonollosa. 2016. Neural machine translation using bitmap fonts. In *Proceedings of the EAMT 2016 Fifth Workshop on Hybrid Approaches to Translation (HyTra)*, pages 1–9.

Loïc Barrault, Magdalena Biesialska, Ondřej Bojar, Marta R. Costa-jussà, Christian Federmann, Yvette Graham, Roman Grundkiewicz, Barry Haddow, Matthias Huck, Eric Joanis, Tom Kocmi, Philipp Koehn, Chi-kiu Lo, Nikola Ljubešić, Christof Monz, Makoto Morishita, Masaaki Nagata, Toshiaki Nakazawa, Santanu Pal, Matt Post, and Marcos Zampieri. 2020. Findings of the 2020 conference on machine translation (WMT20). In *Proceedings of the Fifth Conference on Machine Translation*, pages 1–55, Online. Association for Computational Linguistics.

Yonatan Belinkov and Yonatan Bisk. 2018. Synthetic and natural noise both break neural machine translation. In *International Conference on Learning Representations*.

Isaac Caswell, Theresa Breiner, Daan van Esch, and Ankur Bapna. 2020. Language ID in the wild: Unexpected challenges on the path to a thousand-language web text corpus. In *Proceedings of the 28th International Conference on Computational Linguistics*, pages 6588–6608, Barcelona, Spain (Online). International Committee on Computational Linguistics.

Colin Cherry, George Foster, Ankur Bapna, Orhan Firat, and Wolfgang Macherey. 2018. Revisiting character-based neural machine translation with capacity and compression. In *Proceedings of the 2018 Conference on Empirical Methods in Natural Language Processing*, pages 4295–4305, Brussels, Belgium. Association for Computational Linguistics.

Jonathan H Clark, Dan Garrette, Iulia Turc, and John Wieting. 2021. Canine: Pre-training an efficient tokenization-free encoder for language representation. *arXiv preprint arXiv:2103.06874*.

Marta R. Costa-jussà, David Aldón, and José A. R. Fonollosa. 2017. Chinese–spanish neural machine translation enhanced with character and word bitmap fonts. *Machine Translation*, 31(1):35–47.

Falcon Dai and Zheng Cai. 2017. Glyph-aware embedding of chinese characters. In *Proceedings of the First Workshop on Subword and Character Level Models in NLP*, pages 64–69.

Shuoyang Ding, Adithya Renduchintala, and Kevin Duh. 2019. A call for prudent choice of subword merge operations in neural machine translation. In *Proceedings of Machine Translation Summit XVII Volume 1: Research Track*, pages 204–213, Dublin, Ireland. European Association for Machine Translation.

Alexey Dosovitskiy, Lucas Beyer, Alexander Kolesnikov, Dirk Weissenborn, Xiaohua Zhai, Thomas Unterthiner, Mostafa Dehghani, Matthias Minderer, Georg Heigold, Sylvain Gelly, Jakob Uszkoreit, and Neil Houlsby. 2021. An image is worth 16x16 words: Transformers for image recognition at scale. In *International Conference on Learning Representations*.

Kevin Duh. 2018. The multitarget TED talks task. `http://www.cs.jhu.edu/~kevinduh/a/multitarget-tedtalks/`.

Steffen Eger, Gözde Gül Şahin, Andreas Rücklé, Ji-Ung Lee, Claudia Schulz, Mohsen Mesgar, Krishnkant Swarnkar, Edwin Simpson, and Iryna Gurevych. 2019. Text processing like humans do: Visually attacking and shielding nlp systems. In *Proceedings of NAACL-HLT*, pages 1634–1647.

Armand Joulin, Edouard Grave, Piotr Bojanowski, Matthijs Douze, Hérve Jégou, and Tomas Mikolov. 2016a. Fasttext.zip: Compressing text classification models. *arXiv preprint arXiv:1612.03651*.

Armand Joulin, Edouard Grave, Piotr Bojanowski, and Tomas Mikolov. 2016b. Bag of tricks for efficient text classification. *arXiv preprint arXiv:1607.01759*.

Marcin Junczys-Dowmunt, Bruno Pouliquen, and Christophe Mazenc. 2016. Coppa v2. 0: Corpus of parallel patent applications building large parallel corpora with gnu make. In *4th Workshop on Challenges in the Management of Large Corpora Workshop Programme*.

Huda Khayrallah and Philipp Koehn. 2018. On the impact of various types of noise on neural machine translation. In *Proceedings of the 2nd Workshop on Neural Machine Translation and Generation*, pages 74–83, Melbourne, Australia. Association for Computational Linguistics.

Yoon Kim, Yacine Jernite, David Sontag, and Alexander M Rush. 2016. Character-aware neural language models. In *Proceedings of the AAAI Conference on Artificial Intelligence*, volume 30, pages 2741–2749.

Taku Kudo. 2018. Subword regularization: Improving neural network translation models with multiple subword candidates. In *Proceedings of the 56th Annual Meeting of the Association for Computational Linguistics (Volume 1: Long Papers)*, pages 66–75.

Xian Li, Paul Michel, Antonios Anastasopoulos, Yonatan Belinkov, Nadir Durrani, Orhan Firat,

Philipp Koehn, Graham Neubig, Juan Pino, and Hassan Sajjad. 2019. Findings of the first shared task on machine translation robustness. In *Proceedings of the Fourth Conference on Machine Translation (Volume 2: Shared Task Papers, Day 1)*, pages 91–102, Florence, Italy. Association for Computational Linguistics.

Frederick Liu, Han Lu, Chieh Lo, and Graham Neubig. 2017. Learning character-level compositionality with visual features. In *Proceedings of the 55th Annual Meeting of the Association for Computational Linguistics (Volume 1: Long Papers)*, pages 2059–2068.

Elman Mansimov, Mitchell Stern, Mia Xu Chen, Orhan Firat, Jakob Uszkoreit, and Puneet Jain. 2020. Towards end-to-end in-image neural machine translation. In *Proceedings of the First International Workshop on Natural Language Processing Beyond Text*, pages 70–74.

Leo X McCusker, Philip B Gough, and Randolph G Bias. 1981. Word recognition inside out and outside in. *Journal of Experimental Psychology: Human Perception and Performance*, 7(3):538.

Yuxian Meng, Wei Wu, Fei Wang, Xiaoya Li, Ping Nie, Fan Yin, Muyu Li, Qinghong Han, Xiaofei Sun, and Jiwei Li. 2019. Glyce: Glyph-vectors for chinese character representations. In *Advances in Neural Information Processing Systems*, volume 32. Curran Associates, Inc.

Paul Michel and Graham Neubig. 2018. MTNT: A testbed for machine translation of noisy text. In *Proceedings of the 2018 Conference on Empirical Methods in Natural Language Processing*, pages 543–553, Brussels, Belgium. Association for Computational Linguistics.

Markus Näther. 2020. An in-depth comparison of 14 spelling correction tools on a common benchmark. In *Proceedings of the 12th Language Resources and Evaluation Conference*, pages 1849–1857, Marseille, France. European Language Resources Association.

Graham Neubig, Yosuke Nakata, and Shinsuke Mori. 2011. Pointwise prediction for robust, adaptable Japanese morphological analysis. In *Proceedings of the 49th Annual Meeting of the Association for Computational Linguistics: Human Language Technologies*, pages 529–533, Portland, Oregon, USA. Association for Computational Linguistics.

Nikola I. Nikolov, Yuhuang Hu, Mi Xue Tan, and Richard H. R. Hahnloser. 2018. Character-level chinese-english translation through ascii encoding.

Ossama Obeid, Nasser Zalmout, Salam Khalifa, Dima Taji, Mai Oudah, Bashar Alhafni, Go Inoue, Fadhl Eryani, Alexander Erdmann, and Nizar Habash. 2020. CAMeL tools: An open source python toolkit for Arabic natural language processing. In *Proceedings of the 12th Language Resources and Evaluation Conference*, pages 7022–7032, Marseille, France. European Language Resources Association.

Myle Ott, Sergey Edunov, Alexei Baevski, Angela Fan, Sam Gross, Nathan Ng, David Grangier, and Michael Auli. 2019. fairseq: A fast, extensible toolkit for sequence modeling. In *Proceedings of NAACL-HLT 2019: Demonstrations*.

Manuel Perea, J. A. Duñabeitia, and M. Carreiras. 2008. R34d1ng w0rd5 w1th numb3r5. *Journal of experimental psychology. Human perception and performance*, 34 1:237–41.

Ivan Provilkov, Dmitrii Emelianenko, and Elena Voita. 2020. Bpe-dropout: Simple and effective subword regularization. In *Proceedings of the 58th Annual Meeting of the Association for Computational Linguistics*, pages 1882–1892.

Alec Radford, Jeff Wu, Rewon Child, David Luan, Dario Amodei, and Ilya Sutskever. 2019. Language models are unsupervised multitask learners.

Stephen Rawls, Huaigu Cao, Senthil Kumar, and Prem Natarjan. 2017. Combining convolutional neural networks and LSTMs for segmentation free OCR. In *Proc. ICDAR*.

K. Rayner, S. White, Rebecca Lynn Johnson, and S. Liversedge. 2006. Raeding wrods with jubmled lettres. *Psychological Science*, 17:192 – 193.

Ricardo Rei, Craig Stewart, Ana C Farinha, and Alon Lavie. 2020. COMET: A neural framework for MT evaluation. In *Proceedings of the 2020 Conference on Empirical Methods in Natural Language Processing (EMNLP)*, pages 2685–2702, Online. Association for Computational Linguistics.

Shruti Rijhwani, Antonios Anastasopoulos, and Graham Neubig. 2020. OCR Post Correction for Endangered Language Texts. In *Proceedings of the 2020 Conference on Empirical Methods in Natural Language Processing (EMNLP)*, pages 5931–5942, Online. Association for Computational Linguistics.

Maria Ryskina, Matthew R. Gormley, and Taylor Berg-Kirkpatrick. 2020. Phonetic and visual priors for decipherment of informal Romanization. In *Proceedings of the 58th Annual Meeting of the Association for Computational Linguistics*, pages 8308–8319, Online. Association for Computational Linguistics.

Keisuke Sakaguchi, Kevin Duh, Matt Post, and Benjamin Van Durme. 2017. Robsut wrod reocginiton via semi-character recurrent neural network. In *Thirty-First AAAI Conference on Artificial Intelligence*.

Elizabeth Salesky, Andrew Runge, Alex Coda, Jan Niehues, and Graham Neubig. 2018. Optimizing segmentation granularity for neural machine translation. *arXiv preprint arXiv:1810.08641*.

Rico Sennrich, Barry Haddow, and Alexandra Birch. 2016. Neural machine translation of rare words with subword units. In *Proceedings of the 54th Annual Meeting of the Association for Computational Linguistics (Volume 1: Long Papers)*, pages 1715–1725, Berlin, Germany. Association for Computational Linguistics.

Pamela Shapiro and Kevin Duh. 2018. BPE and CharCNNs for translation of morphology: A cross-lingual comparison and analysis. *arXiv preprint arXiv:1809.01301*.

Lucia Specia, Zhenhao Li, Juan Pino, Vishrav Chaudhary, Francisco Guzmán, Paul Michel, Graham Neubig, Hassan Sajjad, Nadir Durrani, Yonatan Belinkov, Philipp Koehn, and Xian Li. 2020. Findings of the WMT 2020 Shared Task on Machine Translation Robustness. In *Proceedings of the Fifth Conference on Machine Translation: Shared Task Papers*.

Baohua Sun, Lin Yang, Catherine Chi, Wenhan Zhang, and Michael Lin. 2019. Squared english word: A method of generating glyph to use super characters for sentiment analysis. In *AffCon@AAAI*.

Baohua Sun, Lin Yang, Patrick Dong, Wenhan Zhang, Jason Dong, and Charles Young. 2018. Super characters: A conversion from sentiment classification to image classification. In *Proceedings of the 9th Workshop on Computational Approaches to Subjectivity, Sentiment and Social Media Analysis*, pages 309–315, Brussels, Belgium. Association for Computational Linguistics.

Vaibhav Vaibhav, Sumeet Singh, Craig Stewart, and Graham Neubig. 2019. Improving robustness of machine translation with synthetic noise. In *Proceedings of the 2019 Conference of the North American Chapter of the Association for Computational Linguistics: Human Language Technologies, Volume 1 (Long and Short Papers)*, pages 1916–1920, Minneapolis, Minnesota. Association for Computational Linguistics.

Changhan Wang, Kyunghyun Cho, and Jiatao Gu. 2020a. Neural machine translation with byte-level subwords. In *Proceedings of the AAAI Conference on Artificial Intelligence*, volume 34, pages 9154–9160.

Haohan Wang, Peiyan Zhang, and Eric P Xing. 2020b. Word shape matters: Robust machine translation with visual embedding. *arXiv preprint arXiv:2010.09997*.

# A Parameter tuning

Additional parameter tuning results by language pair for MTTT; Table 9 results for DE-EN can be found in the main text (Table 5). Window length is always greater than or equal to stride length so that no text is dropped. With longer sequence lengths (shorter strides) and smaller batch sizes, we observe occasional instability (similar to character-based text models), which increased batch sizes generally stabilize.
$w$ = window size, $s$ = stride, $c$ = number of convolutional blocks.

Table 8: Translation results for MTTT, tuning stride and window length with fixed batch size $20k$ and font size 10.

| AR–EN | $c=1,\ font=10pt$ | | | | | | |
|---|---|---|---|---|---|---|---|
| $s\downarrow/w\rightarrow$ | 10 | 15 | 20 | 25 | 30 | 35 | 40 |
| 5 | 30.8 | 30.0 | 30.6 | 31.3 | 30.4 | 30.5 | 29.6 |
| 10 | 30.3 | 28.5 | 31.0 | 31.6 | 31.4 | 31.4 | 30.4 |
| 15 | | 25.2 | 30.2 | 30.3 | 30.6 | 29.4 | 29.3 |

| DE–EN | $c=1,\ font=10pt$ | | | | | | |
|---|---|---|---|---|---|---|---|
| $s\downarrow/w\rightarrow$ | 10 | 15 | 20 | 25 | 30 | 35 | 40 |
| 5 | 0.7 | 32.6 | 35.1 | 0.5 | 33.1 | 33.9 | 32.5 |
| 10 | 0.6 | 34.6 | 34.8 | 32.8 | 32.9 | 34.4 | 33.5 |
| 15 | | 32.8 | 33.9 | 32.0 | 31.4 | 33.7 | 33.9 |

| FR–EN | $c=1,\ font=10pt$ | | | | | | |
|---|---|---|---|---|---|---|---|
| $s\downarrow/w\rightarrow$ | 10 | 15 | 20 | 25 | 30 | 35 | 40 |
| 5 | 35.4 | 35.7 | 35.7 | 35.5 | 0.7 | 0.6 | 0.8 |
| 10 | 35.6 | 36.2 | 36.1 | 36.1 | 34.7 | 34.7 | 35.0 |
| 15 | | 35.7 | 35.8 | 35.6 | 34.4 | 34.3 | 34.6 |

| JA–EN | $c=1,\ font=10pt$ | | | | | | |
|---|---|---|---|---|---|---|---|
| $s\downarrow/w\rightarrow$ | 10 | 15 | 20 | 25 | 30 | 35 | 40 |
| 5 | 12.4 | 11.5 | 12.3 | 13.1 | 12.4 | 12.4 | 12.3 |
| 10 | 11.8 | 11.8 | 12.4 | 12.5 | 11.5 | 12.4 | 12.3 |
| 15 | | 9.4 | 12.1 | 12.7 | 12.2 | 12.4 | 12.1 |

| KO–EN | $c=1,\ font=10pt$ | | | | | | |
|---|---|---|---|---|---|---|---|
| $s\downarrow/w\rightarrow$ | 10 | 15 | 20 | 25 | 30 | 35 | 40 |
| 5 | 15.8 | 15.7 | 15.3 | 16.2 | 15.6 | 16.0 | 16.1 |
| 10 | 14.7 | 15.9 | 15.5 | 16.5 | 14.7 | 15.9 | 16.4 |
| 15 | | 14.3 | 15.2 | 15.4 | 15.7 | 16.2 | 15.6 |

| RU–EN | $c=1,\ font=10pt$ | | | | | | |
|---|---|---|---|---|---|---|---|
| $s\downarrow/w\rightarrow$ | 10 | 15 | 20 | 25 | 30 | 35 | 40 |
| 5 | 0.6 | 22.7 | 23.8 | 0.5 | 23.6 | 23.0 | 0.5 |
| 10 | 2.0 | 23.2 | 25.0 | 23.2 | 23.2 | 23.9 | 23.2 |
| 15 | | 21.1 | 24.4 | 23.7 | 24.5 | 24.2 | 22.0 |

| ZH–EN | $c=1,\ font=10pt$ | | | | | | |
|---|---|---|---|---|---|---|---|
| $s\downarrow/w\rightarrow$ | 10 | 15 | 20 | 25 | 30 | 35 | 40 |
| 5 | 16.7 | 0.4 | 16.7 | 17.3 | 17.4 | 17.0 | 0.4 |
| 10 | 15.8 | 17.1 | 16.8 | 17.1 | 16.3 | 17.0 | 0.4 |
| 15 | | 16.0 | 16.0 | 16.3 | 16.4 | 16.3 | 0.5 |

Varied conv. kernel size (note: 23=full window height).

| $h \times w$ | $3 \times 3$ | $3 \times 1$ | $1 \times 3$ | $13 \times 3$ | $23 \times 3$ | $5 \times 5$ |
|---|---|---|---|---|---|---|
| ZH–EN | 17.4 | 17.1 | 16.9 | 16.7 | 0.6 | 16.6 |

Table 9: Translation results for MTTT, tuning stride and window length with fixed batch size $10k$ and font size 8.

| AR–EN | $c=1,\ font=8pt$ | | | |
|---|---|---|---|---|
| $s\downarrow/w\rightarrow$ | 15 | 20 | 25 | 30 |
| 10 | 29.5 | 29.7 | 27.1 | 30.0 |
| 15 | 27.2 | 28.5 | 24.0 | 28.6 |
| 20 | | 26.0 | 11.5 | 26.9 |
| 25 | | | 19.9 | 25.4 |

| DE–EN | $c=1,\ font=8pt$ | | | |
|---|---|---|---|---|
| $s\downarrow/w\rightarrow$ | 15 | 20 | 25 | 30 |
| 10 | 33.4 | 33.1 | 33.3 | 33.6 |
| 15 | 33.9 | 32.9 | 31.3 | 32.9 |
| 20 | | 32.0 | 30.3 | 32.4 |
| 25 | | | 30.4 | 30.9 |

| FR–EN | $c=1,\ font=8pt$ | | | |
|---|---|---|---|---|
| $s\downarrow/w\rightarrow$ | 15 | 20 | 25 | 30 |
| 10 | 33.9 | 35.1 | 35.5 | 35.0 |
| 15 | 34.8 | 34.5 | 34.8 | 33.5 |
| 20 | | 33.6 | 34.0 | 33.5 |
| 25 | | | 33.7 | 32.6 |

| JA–EN | $c=1,\ font=8pt$ | | | |
|---|---|---|---|---|
| $s\downarrow/w\rightarrow$ | 15 | 20 | 25 | 30 |
| 10 | 10.7 | 11.5 | 10.9 | 11.1 |
| 15 | 8.7 | 10.5 | 10.9 | 10.0 |
| 20 | | 10.3 | 8.2 | 8.9 |
| 25 | | | 8.3 | 8.0 |

| KO–EN | $c=1,\ font=8pt$ | | | |
|---|---|---|---|---|
| $s\downarrow/w\rightarrow$ | 15 | 20 | 25 | 30 |
| 10 | 15.2 | 14.8 | 14.7 | 14.6 |
| 15 | 14.7 | 14.9 | 14.3 | 14.3 |
| 20 | | 13.9 | 14.1 | 13.3 |
| 25 | | | 13.5 | 12.3 |

| RU–EN | $c=1,\ font=8pt$ | | | |
|---|---|---|---|---|
| $s\downarrow/w\rightarrow$ | 15 | 20 | 25 | 30 |
| 10 | 19.6 | 0.6 | 23.8 | 23.3 |
| 15 | 22.4 | 0.5 | 22.8 | 23.2 |
| 20 | | 0.6 | 22.3 | 22.5 |
| 25 | | | 21.9 | 21.4 |

| ZH–EN | $c=1,\ font=8pt$ | | | |
|---|---|---|---|---|
| $s\downarrow/w\rightarrow$ | 15 | 20 | 25 | 30 |
| 10 | 0.5 | 14.4 | 0.5 | 0.5 |
| 15 | 12.0 | 13.4 | 0.6 | 0.6 |
| 20 | | 12.3 | 0.5 | 4.3 |
| 25 | | | 0.6 | 5.3 |

## B Ablation experiments

To what degree are our results due to our implicit model of segmentation through overlapping sliding windows, or the use of visual representations themselves? To disentangle these two factors, we run ablation experiments to separate these two components of the visual embedder.

**Sliding window segmentation only.** To evaluate our approach to segmentation without visual rendering, we apply sliding window vocabularies to text, creating overlapping character 3-grams: this corresponds to a window size of approximately 30 with font size 10. Character n-grams of a fixed order are not commonly used for NMT, likely due to the large resulting vocabulary and the fact that they do not solve the OOV problem.

For languages with (more) uniform character n-gram frequencies (*Arabic, German, French, Russian*), results with sliding window segmentation but no visual representations (*w/o visrep*, char n-grams ablation) are similar to the text BPE models' results. For these four languages, the sliding window approach to segmentation does not affect performance (and for German, the sliding windows in fact provide a +1 BLEU improvement over BPE). For French, we see slight degradation *with* the visual representations compared to the ablation (0.2 BLEU), suggesting that the visual embedder itself has slight room for improvement. For the other languages (*Chinese, Japanese, Korean*), there is a significant drop *w/o visrep* due to the higher proportion of infrequently observed vocabulary, leading to a greater proportion of insufficiently trained embeddings. This is a problem that our visual text embedder removes in the full visual text models, because exact lexical matches are not required to train visual representations.

| `MODEL:` | **ar** | **de** | **fr** | **ja** | **ko** | **ru** | **zh** |
|---|---|---|---|---|---|---|---|
| Visual text | 31.6 | 35.1 | 36.2 | 13.1 | 16.6 | 25.0 | 17.6 |
| *w/o visrep* (char n-grams) | 31.5 | 34.6 | 36.4 | 1.4 | 1.3 | 24.6 | 5.5 |
| Text, BPE | 32.1 | 33.6 | 36.7 | 14.4 | 17.0 | 25.4 | 18.3 |
| `NOISED:` | | | | | | | |
| Visual text; `swap p=0.5` | 21.7 | 29.4 | 28.4 | — | 11.5 | 18.3 | — |
| *w/o visrep*; `swap p=0.5` | 11.2 | 10.8 | 11.9 | — | 1.1 | 9.5 | — |
| Text, BPE; `swap p=0.5` | 12.4 | 13.1 | 13.3 | — | 10.8 | 11.1 | — |

Table 10: **Ablation:** Sliding window segmentation (character n-grams) applied to text without visual rendering.

When we add noise, the ablation experiments (sliding window segmentation without visual representations, *w/o visrep*) degrade below the BPE baselines; this suggests that the visual text embedder (combined with the resulting open vocabulary) is the primary reason for our visual text models' robustness, not our sliding window segmentation.

## C Normalization as preprocessing for robustness

A natural question is whether preprocessing can address the robustness issues demonstrated here with traditional text models using e.g., BPE subword segmentation. To evaluate this setting, we apply a spellchecker to each of our noise-induced test sets; for this task, we use the test sets with noise induced with $p = 1.0$, where 100% of applicable tokens have induced noise. We use the Google Docs spellchecker, which was the best of the options evaluated in recent work (Näther, 2020) which cover all of our tested languages (unlike e.g., Grammarly, which currently supports English only), and which significantly outperformed common open-source alternatives such as Hunspell.[12] We evaluate both the text *(BPE)* and visual text *(visrep)* models on the spellchecked test sets; results are shown in Table 11.

It is clear that spellchecking can help the BPE models, in some cases dramatically (up to 20 BLEU). However, it does not close the gap with our method, and in some cases performance degrades; for all induced noise, visual text representations still outperform the BPE models, and often by a large margin.

Spellcheckers are language-specific and as shown below in Table 11, can be more adept at certain types of noise which were taken into consideration in their construction. For example, while first spellchecking the French `swap` test set improves the *BPE* model by more than 20 BLEU, it does not change the `l33tspeak` performance at all. Similarly, *BPE* models were only slightly improved for Arabic diacritization and Russian unicode noise, while the *visrep* model performs strongly for both without spellcheck. Further, like translation models, spellcheckers often rely on context for disambiguation, and so with noisy context may either have lower recall or can introduce cascading errors when the correction made is not correct (illustrated below in *lower* performance for some conditions with spellcheck). A denoising autoencoder may also be able to address many of these phenomena, but, requires training and knowledge of the types of noise expected, where our approach is a single model and performance is zero-shot. Possible noise grows exponentially as it can appear in combination — it is not feasible to expect normalization to fully address this problem.

| | | **Arabic** | | **French** | | **German** | | **Korean** | | **Russian** | |
|---|---|---|---|---|---|---|---|---|---|---|---|
| | | *BPE* | *visrep* | *BPE* | *visrep* | *BPE* | *visrep* | *BPE* | *visrep* | *BPE* | *visrep* |
| | no noise | **32.1** | 31.6 | **36.7** | 36.2 | 33.6 | **35.1** | **17.0** | 16.6 | **25.4** | 25.0 |
| **swap** | induced noise | 2.3 | **9.3** | 2.4 | **22.0** | 1.9 | **25.9** | 5.4 | **8.9** | 5.4 | **18.8** |
| | + *spellcheck* | 7.9 | **11.9** | 23.8 | **29.1** | 1.9 | **14.1** | 5.1 | **6.9** | 10.8 | **18.2** |
| **cambridge** | induced noise | 7.8 | **13.2** | 6.9 | **18.3** | 6.5 | **16.9** | 12.6 | **14.1** | 4.5 | **11.1** |
| | + *spellcheck* | 10.9 | **12.6** | 16.4 | **21.1** | 10.0 | **14.9** | 10.3 | **11.8** | 5.9 | **11.1** |
| **l33tspeak** | induced noise | — | — | 0.3 | **0.7** | 0.7 | **1.2** | — | — | — | — |
| | + *spellcheck* | — | — | 0.3 | **0.7** | 0.7 | **1.2** | — | — | — | — |
| **diacritics** | induced noise | 1.7 | **25.2** | — | — | — | — | — | — | — | — |
| | + *spellcheck* | 2.1 | **25.3** | — | — | — | — | — | — | — | — |
| **unicode** | induced noise | — | — | — | — | — | — | — | — | 1.6 | **22.0** |
| | + *spellcheck* | — | — | — | — | — | — | — | — | 2.1 | **20.4** |

Table 11: Translation performance on five types of induced noise with spellchecking as preprocessing; all test sets have noise induced with $p = 1.0$. Both traditional text models *(BPE)* and visual text models *(visrep)* are shown. We bold the best model for each condition.

[12] https://github.com/hunspell/hunspell

## D Additional robustness figures

Here we show character permutation results isolated by model and noise type, and absolute BLEU for figures shown with ΔBLEU for readability in the main text.

### D.1 Isolated character permutation results

Each plot in Figure 7 shows the degradation in performance of a given model with different proportions of induced noise, relative to the performance of the *same model* on the uncorrupted text. As more noise is added, the visual text models degrade at significantly lower pace.

Average number of tokens per sentence and average token length affect the amount of noise; for `cmabirdge (cam)` Korean appears to be an outlier because there are fewer words where this noise may be applied than our other languages, as there are fewer words of length $\geq 4$ in the data.

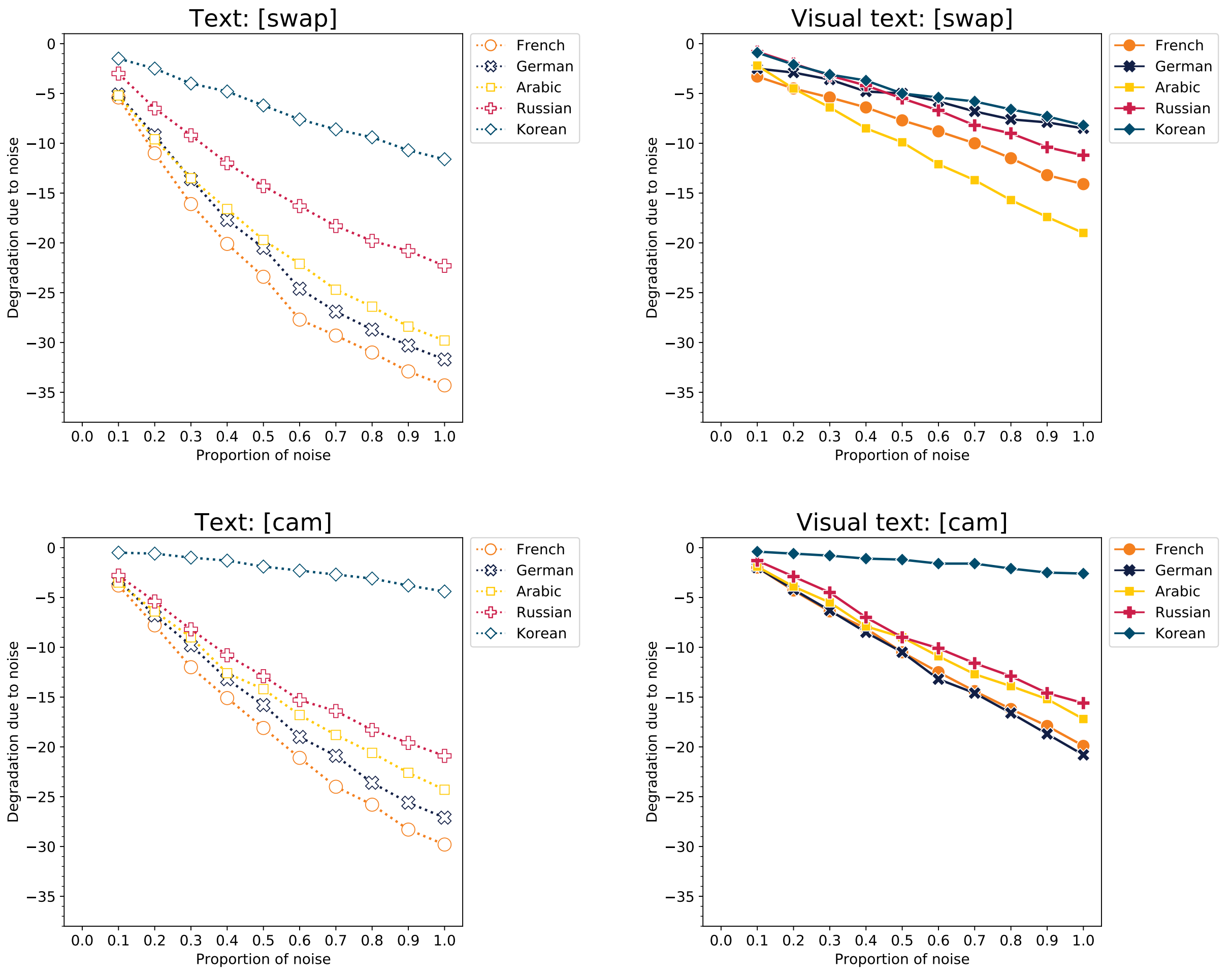

Figure 7: **Degradation due to noise in the form of character permutations**. Each point represents how much worse that model does with a given proportion of noise, relative to the *same model* on uncorrupted text.
[Top] **`swap`** of two characters within a token. [Bottom] **`cmabirdge`** word-internal permutations (cam);
[Left] Text model baselines; [Right] Visual text models.

## D.2 Absolute BLEU

Here we show absolute BLEU for figures shown with ΔBLEU for readability in the main text.

Average number of tokens per sentence and average token length affect the amount of noise; for `cmabirdge (cam)` Korean appears to be an outlier because there are fewer words where this noise may be applied than our other languages, as there are fewer words of length $\geq 4$ in the data.

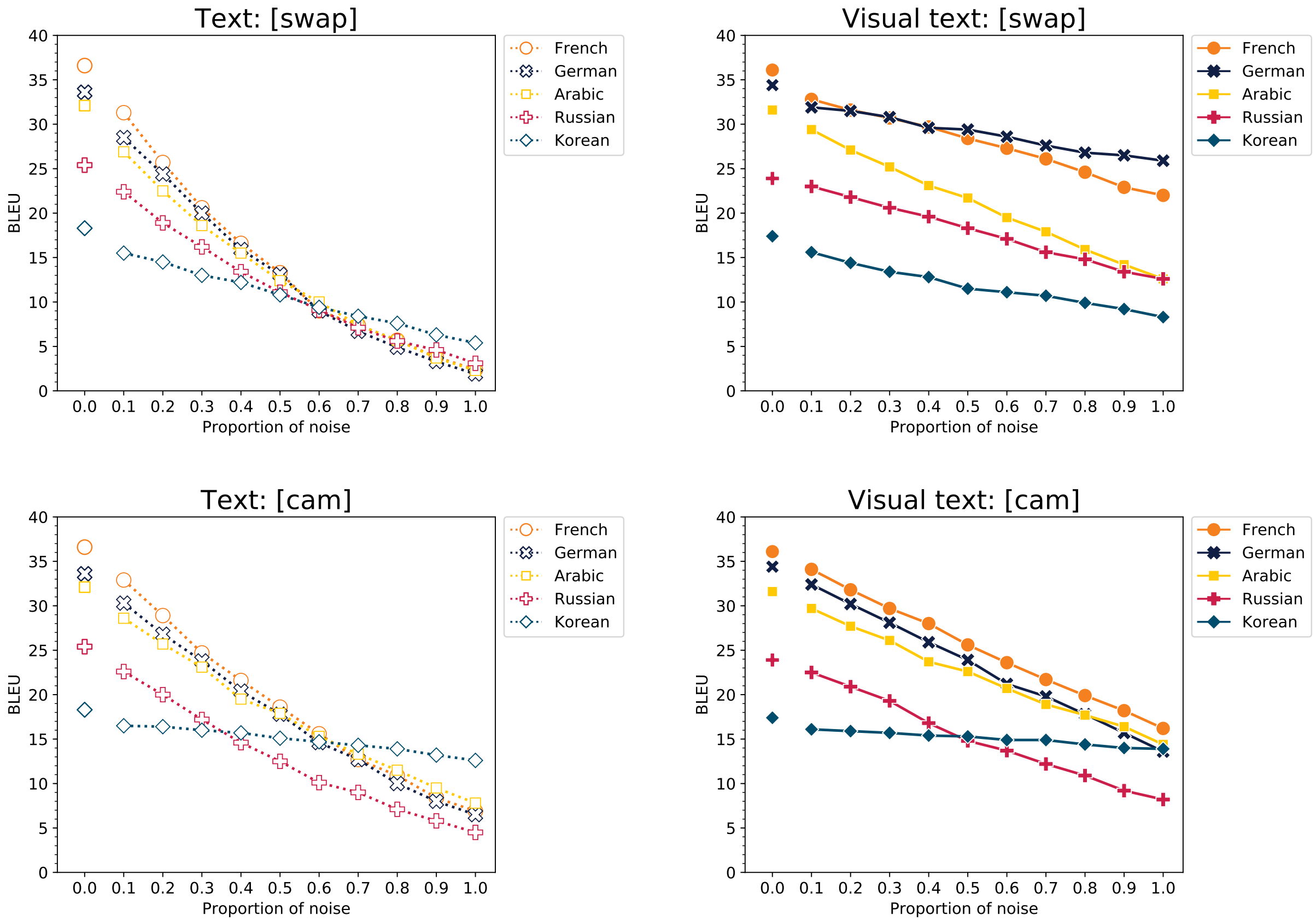

Figure 8: **Character permutations.** Absolute BLEU is shown. See Figure 6 for ΔBLEU for readability. [Top] `swap` of two characters within a token. [Bottom] `cmabirdge` word-internal permutations (cam); [Left] Text model baselines; [Right] Visual text models.

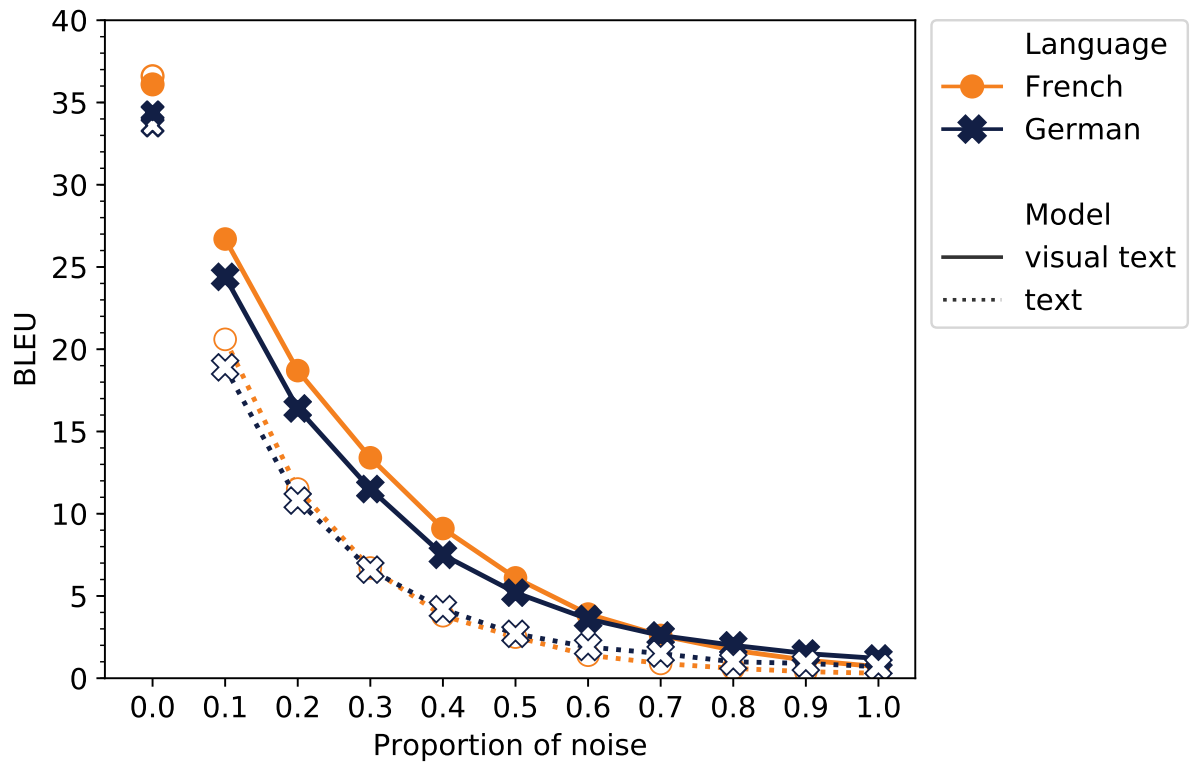

Figure 9: **Visual noise: `l33tspeak`.** Absolute BLEU is shown. See Figure 5 for ΔBLEU for readability. For `l33tspeak`, improvements with visual text diminish with higher levels of noise.

## E Pixel Density

Below we show the average pixel density (the average pixel value, normalized to be between 0 and 1, where 0 is white and 1 is black) and percentage of non-white pixels for rendered text. We find that pixel density is not necessarily indicative of performance, but that those languages with lower pixel densities are less sensitive to differences in font size (see parameter grids in Appendix A). Arabic diacritization yields an approximately 2% increase in pixel density.

| | **ar** | **de** | **fr** | **ja** | **ko** | **ru** | **zh** |
|---|---|---|---|---|---|---|---|
| Avg. pixel density, *fontsize=10pt* | 0.08 | 0.14 | 0.14 | 0.14 | 0.15 | 0.14 | 0.17 |
| Avg. pixel density, *fontsize=8pt* | 0.07 | 0.13 | 0.13 | 0.13 | 0.14 | 0.13 | 0.17 |
| % non-white pixels, *fontsize=10pt* | 0.17 | 0.24 | 0.24 | 0.25 | 0.28 | 0.24 | 0.32 |
| % non-white pixels, *fontsize=8pt* | 0.21 | 0.24 | 0.24 | 0.26 | 0.25 | 0.24 | 0.31 |

Table 12: **Pixel density information:** shown across our 7 languages for rendered text at font sizes 8pt and 10pt.

## F WMT Data

Table 13 contains further information about the data used to train the larger WMT models.

| **Task** | **Source** | **Words** | **Segments** |
|---|---|---|---|
| German–English | Europarl v10 | 97,695,640 | 1,828,521 |
| | New Commentary v15 | 16,078,167 | 371,225 |
| | Wikititles v2 | 6,232,820 | 1,382,687 |
| | Tilde Rapid | 46,404,479 | 1,631,639 |
| | WikiMatrix | 56,216,773 | 1,573,438 |
| | TOTAL | 520,776,198 | 5,247,778 |
| | after filtering | 389,457,370 | 4,965,008 |
| Chinese–English | News Commentary v15 | 7,451,529 | 320,713 |
| | Wikititles v2 | 3,013,119 | 836,683 |
| | CCMT | 192,702,068 | 9,023,456 |
| | WikiMatrix | 18,266,293 | 786,512 |
| | TOTAL | 440,691,851 | 10,516,339 |
| | after filtering | 440,917,047 | 8,742,057 |

Table 13: Data used in training the larger WMT systems. **Words** denotes total tokens over both sides, while **Segments** counts parallel sentence pairs. Counts are computed from raw data prior to any filtering.

## G Subword Regularization and BPE-Dropout

Subword regularization is a technique where different subword segmentations are used during each epoch in training, often providing improved performance and robustness. The idea was introduced by Kudo (2018) with a unigram segmentation model that provided the ability to sample multiple candidate segmentations. Inspired by its success, Provilkov et al. (2020) proposed BPE dropout, which achieves a similar effect by randomly dropping operations from the deterministic BPE merge operation sequence. These techniques have been shown to increase model robustness and could therefore improve our text baselines. We therefore compare both methods with visual text here to see how the robustness improvements compare.

We follow the recommended settings from each paper: subword regularization with $l = 64, \alpha = 0.1$ and BPE-dropout with $p = 0.1$. We apply both methods to the source only and do not apply regularization at inference time, and use the vocabulary sizes determined by our tuning in Section 3.3. Subword regularization can yield OOVs when subsequently-sampled subwords are not contained in the model vocabulary; following Provilkov et al. we do not enumerate and include additional subword candidates in the model vocabulary.

**Results on clean text.** Even with our strong results on MTTT, both subword regularization methods yielded improvements of up to 2 BLEU over BPE for some language pairs, with BPE-dropout typically more beneficial. While three language pairs had significant positive improvements (**de, fr, ru**), two had more modest improvements (**ja, ko**) perhaps because many alternate representations were mixtures of characters with subwords, and two were largely unchanged from BPE (**ar, zh**).

| MODEL: | ar | de | fr | ja | ko | ru | zh |
|---|---|---|---|---|---|---|---|
| Visual text | 31.6 | 35.1 | 36.2 | 13.1 | 16.6 | 25.0 | 17.6 |
| Text, BPE | 32.1 | 33.6 | 36.7 | 14.4 | 17.0 | 25.4 | 18.3 |
| Text, Subword Reg. | 32.1 | 36.0 | 37.4 | 14.7 | 17.6 | 26.4 | 18.4 |
| Text, BPE Dropout | 32.3 | 36.7 | 38.1 | 14.8 | 17.8 | 26.7 | 18.3 |

Table 14: Results comparing visual text and BPE models to the two subword regularization techniques on MTTT.

**Results with induced noise.** The primary motivation for this comparison is to see whether subword regularization provides similar model robustness improvements to visual text. We compare both methods to our text and visual text models on the five types of induced noise from Section 5.1.

Figure 10 shows results with `unicode` noise for Russian and `diacritization` for Arabic, adding subword regularization and BPE-dropout performance to Figure 4. Subword regularization provides typical improvements of 4 BLEU over BPE, which BPE-dropout further improves upon, with greater improvement for diacritization than unicode noise (insertions rather than substitutions). However, we see that neither method matches the robustness of visual text. Improved base model performance results in better results than visual text with no noise and noise applied with probability $p = 0.1$, but then a gap quickly emerges as both regularization methods continue to degrade while visual text does not or degrades only slightly.

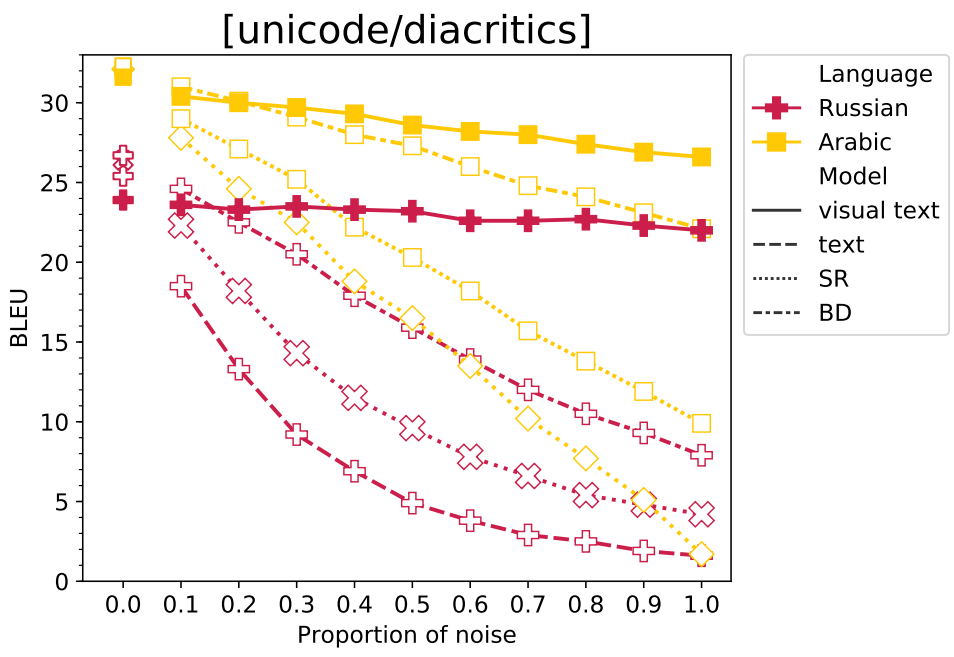

Figure 10: **Visual noise**. Subword regularization (SR) and BPE-dropout (BD) shown with visual text and BPE text models.

In some ways, subword regularization techniques can be viewed similarly to data augmentation. Multiple forms of each token are used throughout training, some of which may better match the subword distribution observed at inference time, while for BPE and visual text only a single segmentation is used—they have no 'adaptation' to noise or variation. Though BPE models may back off to characters faced with unknown words or noise, which are theoretically in vocabulary, character representations have not been observed and in enough contexts in training for the translation model to robustly use them at inference time. With both subword regularization techniques, multiple segmentations which often include characters have been observed in training sufficiently often that the segmentations with noise at inference time can be translated more successfully.

Performance with subword regularization methods is not the same across different types of noise: though both methods generate large improvements for `unicode` noise and `diacritization`, they do not consistently improve over BPE on `l33t`, with average improvements of +0.2 for French and +0.5 for German for BPE-dropout despite improvements of +1.4 and +3.1 without noise.

With character permutations, we see larger improvements on `swap` than `cambridge` with both visual text and subword regularization compared to BPE. We show the results for `swap` in Figure 11. Both methods typically reduce the improvements we see with visual text models (scale: ΔBLEU), but visual text nonetheless provides large improvements over these stronger text models, particularly for French and German—interestingly, these are two of the languages where subword regularization and BPE-dropout provide the largest improvements on clean text and so have the largest performance gap over visual text without any noise. For Korean `swap`, visual text and both regularization methods provide similar improvements. BPE-dropout provides much greater robustness than subword regularization for this type of noise, perhaps because it results in more use of character representations in training, which best matches the subwords which result with this type of noise at inference time. As above, the stronger base performance of the subword regularization methods mean that improvements with visual text are seen with noise of $p > 0.1$.

Future work could investigate if these techniques could be combined, to e.g., use visual text with different image window sizes throughout training to see greater model robustness. Subword regularization and BPE-dropout provide improvements over BPE on both clean text and source noise, making them generally stronger text models. Visual text representations can provide greater robustness still without regularization or adaptation to noise, and on more varied types of noise, making them a promising approach which can likely be improved through combination with complimentary training techniques.

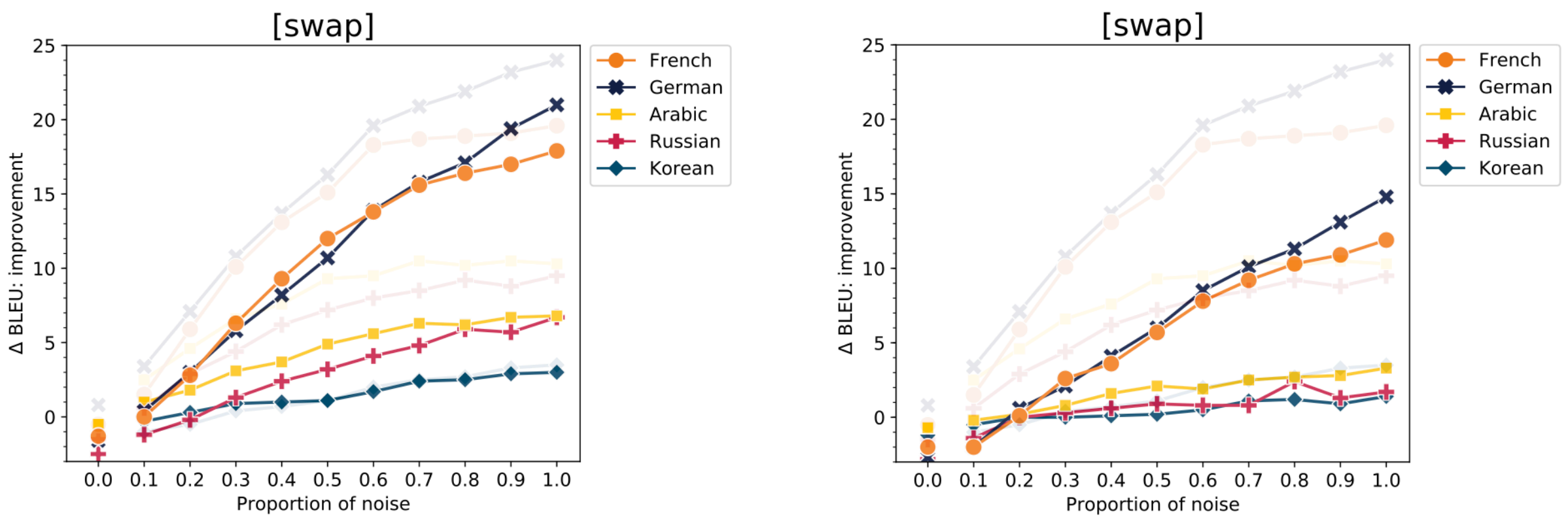

(a) **Subword Regularization**

(b) **BPE-Dropout**

Figure 11: **Character permutations: `swap`.** Visual text improvements over stronger subword regularization baselines are shown in the foreground, with the improvements over BPE shown in the background for context.